\documentclass[letterpaper, 10 pt, conference]{ieeeconf}  

\IEEEoverridecommandlockouts                              

\usepackage{graphics} 
\usepackage{epsfig} 
\usepackage{mathptmx} 
\usepackage{times} 
\usepackage{amsmath} 
\usepackage{amssymb}  
\usepackage{booktabs}
\usepackage{graphicx}
\usepackage{array}
\usepackage{cite}

\makeatletter
\let\NAT@parse\undefined
\makeatother
\usepackage[pagebackref,breaklinks,colorlinks]{hyperref}
\usepackage[capitalize]{cleveref}
\crefname{section}{Sec.}{Secs.}
\Crefname{section}{Section}{Sections}
\Crefname{table}{Table}{Tables}
\crefname{table}{Tab.}{Tabs.}

\begin{document}

\newcommand{\abbrev}{MFSeg }

\title{MFSeg: Efficient Multi-frame 3D Semantic Segmentation}

\author{Chengjie Huang$^{1}$ and Krzysztof Czarnecki$^{2}$
\thanks{$^{1}$Cheriton School of Computer Science, University of Waterloo
        {\tt\small c.huang@uwaterloo.ca}}%
\thanks{$^{2}$Department of Electrical and Computer Engineering, University of Waterloo
        {\tt\small k2czarne@uwaterloo.ca}}%
}
\maketitle

\begin{abstract}
We propose MFSeg, an efficient multi-frame 3D semantic segmentation framework. By aggregating point cloud sequences at the feature level and regularizing the feature extraction and aggregation process, MFSeg reduces computational overhead while maintaining high accuracy. Moreover, by employing a lightweight MLP-based point decoder, our method eliminates the need to upsample redundant points from past frames. Experiments on the nuScenes and Waymo datasets show that MFSeg outperforms existing methods, demonstrating its effectiveness and efficiency.
\end{abstract}

\section{Introduction}
LiDAR-based 3D semantic segmentation is an essential task for 3D scene understanding in robotic applications such as autonomous driving. Accurate segmentation of LiDAR point clouds allows robotic systems to identify and classify objects in their environment, ensuring safe and reliable navigation.

LiDAR point clouds are typically captured sequentially, and utilizing this temporal nature has been shown to enhance various 3D perception tasks. In the context of 3D semantic segmentation, recent work has explored feature-level fusion across consecutive frames~\cite{shi2020spsequencenet,wang2022meta,schutt2022temporallatticenet}. While these approaches have shown promise, they rely on structured representations of point clouds such as range image to align intermediate features across frames or are extensions of older architectures. This limits their applicability to the current state-of-the-art (SOTA) methods in 3D perception such as point transformers~\cite{zhao2021pt,wu2022ptv2,wu2024ptv3,lai2023sphereformer}.

\begin{figure}[ht]
    \centering
    \includegraphics[width=0.95\linewidth]{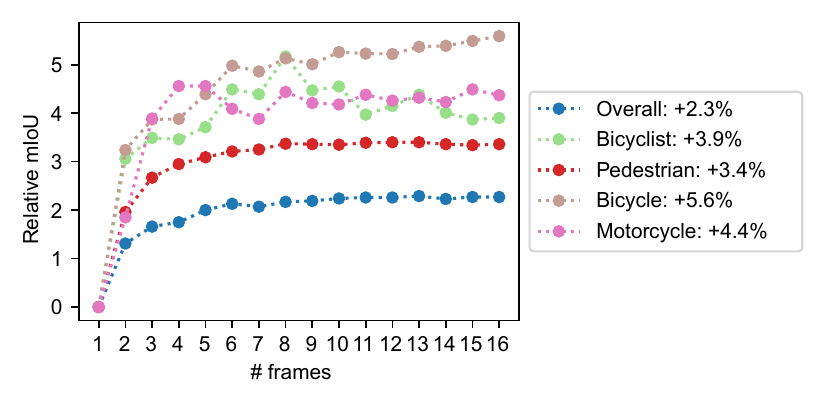}
    \caption{Relative mIoU improvement between various levels of multi-frame point cloud concatenation and single-frame input. Performance are evaluated using a SparseUNet~\cite{graham2018spunet} on the Waymo dataset~\cite{sun2020waymo}.}
    \label{fig:motivation}
\end{figure}

An alternative approach involves aggregating multi-frame point clouds at the input level via concatenation. This is widely adopted in 3D object detection as the multi-frame point clouds contain richer geometric information and can improve detection accuracy~\cite{caesar2020nuscenes,yang20213dman,chen2022mppnet,huang2024soap}. In 3D semantic segmentation, we find this approach to be also beneficial. As shown in \cref{fig:motivation}, multi-frame point clouds provide a modest overall improvement (2.3\%) in mean Intersection over Union (mIoU) performance, and a noticeably higher gain (up to 5.6\%) in detecting vulnerable road users such as pedestrians and cyclists. This highlights the benefits of multi-frame point clouds for 3D semantic segmentation task in real-world applications.

\begin{figure}[ht]
    \centering
    \includegraphics[width=0.49\linewidth]{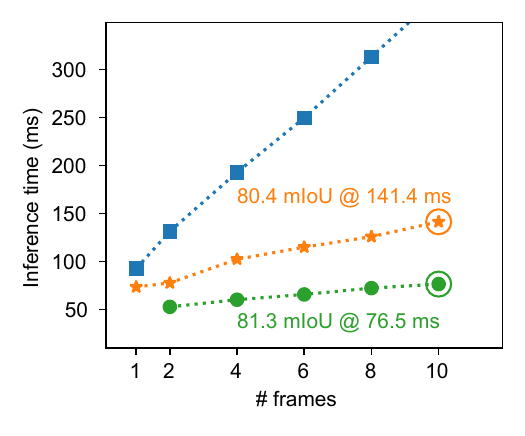}
    \includegraphics[width=0.49\linewidth]{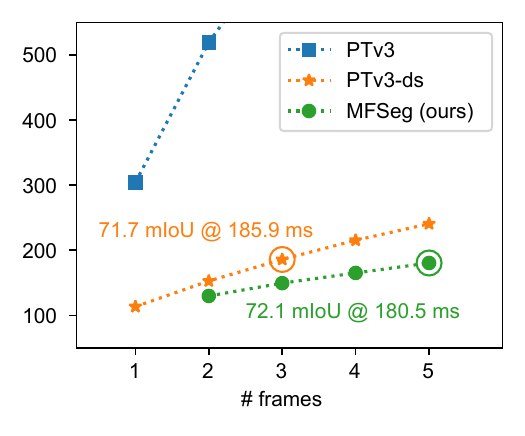}
    \caption{Inference time of PTv3~\cite{wu2024ptv3} and MFSeg under a range of point cloud sequence length on the nuScenes dataset~\cite{caesar2020nuscenes} (left) and Waymo dataset~\cite{sun2020waymo} (right). PTv3-ds denotes that the point cloud is downsampled using a voxel-grid with size 0.05\,$\mathrm{m}^3$, while PTv3 is evaluated with minimum input downsampling.}
    \label{fig:latency}
\end{figure}

Despite the benefits, multi-frame point clouds pose unique challenges for the SOTA point-based methods. Firstly, the increase in the number of points introduces computational challenges. For instance, a 64-beam LiDAR sensor can produce an average of 177k points per frame~\cite{sun2020waymo}, and concatenating six frames can result in over one million points. Processing these dense point clouds requires significant computation, as illustrated by the latency of a SOTA method PTv3~\cite{wu2024ptv3} in \cref{fig:latency}. Secondly, current methods follow the U-Net design and require upsampling point features in the decoder stage. This introduces unnecessary computational overhead for multi-frame point clouds, as predictions are typically only needed for points in the current frame.

In this work, we propose MFSeg, an efficient multi-frame 3D semantic segmentation framework. MFSeg aggregates point cloud sequences at the feature level while retaining the properties of input-level concatenation via a novel aggregation pipeline and regularization strategy. Unlike prior work, our method does not assume a structured feature representation and thus can be applied to SOTA point-based backbones such as point transformers.

To address the computational challenges, we employ a local feature extractor to reduce the input density, allowing MFSeg to efficiently process multi-frame inputs and features. Furthermore, we introduce a simple MLP-based point decoder to replace the traditional symmetrical decoder in the U-Net design, avoiding the unnecessary upsampling of point features. 

These components enable MFSeg to achieve SOTA performance with low latency, as illustrated in \cref{fig:latency}. Experiments on two large-scale semantic segmentation datasets, nuScenes~\cite{caesar2020nuscenes} and Waymo~\cite{sun2020waymo}, demonstrate that MFSeg can outperform existing methods.

\section{Related Work}
3D semantic segmentation methods typically fall into three categories: projection-based, voxel-based, and point-based, depending on the point cloud representation that is being used. Projection-based methods~\cite{wu2018squeezeseg,wu2019squeezesegv2,milioto2019rangenet++,xu2020squeezesegv3,zhang2020polarnet} use either range-image or bird's-eye view (BEV) projections of point clouds, allowing traditional 2D convolutional neural networks (CNNs) to be used.  However, these representations sacrifice the detailed 3D geometric information, limiting their performance. Voxel-based methods~\cite{choy2019minkunet,graham2018spunet} voxelize the point clouds into a structured representation, allowing 3D CNNs to be applied. Although 3D CNNs are computationally demanding, the development of sparse convolutional networks~\cite{graham2017spconv} has greatly improved their efficiency. Point-based methods directly process unstructured point clouds without the need for voxelization or projection. Early works such as PointNet~\cite{qi2017pointnet} and PointNet++~\cite{qi2017pointnet++} laid the foundation for this approach, but their high computational cost limited their application in large outdoor scenes. With the recent advent of efficient transformers~\cite{dao2022flashattention} and the introduction of transformer-based segmentation methods~\cite{zhao2021pt,wu2022ptv2,wu2024ptv3,lai2023sphereformer}, point-based methods have become the current SOTA.

Many approaches have sought to improve 3D segmentation performance by leveraging additional sources of information. One line of work focuses on multi-view fusion, which involves combining multiple representations of the point cloud~\cite{liu2019spvnas,zhu2021cylinder3d,xu2021rpvnet} or integrating other sensory modalities~\cite{yan20222dpass,liu2023uniseg,li2023mseg3d} to create richer feature representations.

Another line of work aims to exploit temporal information by aggregating sequential point cloud data. Using the range-image representation, SpSequenceNet~\cite{shi2020spsequencenet} proposes a cross-frame global attention and local interpolation to fuse CNN features from the previous frame. Meta-RangeSeg~\cite{wang2022meta} instead computes a residual range image representation to capture the temporal difference between frames. However, range image in the multi-frame context can suffer from misalignment due to ego movement. Alternatively, TemporalLatticeNet~\cite{schutt2022temporallatticenet} extends LatticeNet~\cite{luo2020latticenet} and fuses temporal features using recurrent neural networks and a novel abstract flow module. MSeg3D~\cite{li2023mseg3d}, while focusing on multi-modality fusion, uses concatenated point clouds for their point cloud branch to improve segmentation performance. 

In contrast, MFSeg uses a novel formulation of feature space aggregation to retain the benefits of concatenated point cloud. Moreover, unlike voxelized or range-image representations, MFSeg does not rely on structured features, which allows it to utilize SOTA point-based backbones.

\section{Method}
\label{sec:method}
\begin{figure}
    \centering
    \includegraphics[width=\linewidth]{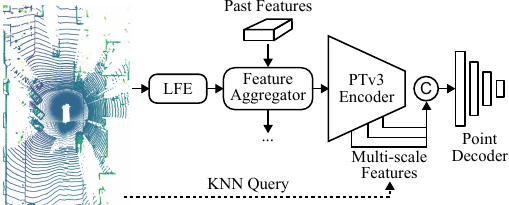}
    \caption{An overview of the proposed \abbrev architecture.}
    \label{fig:architecture}
\end{figure}

\abbrev consists of four key components: (i) a local feature extractor (LFE) for capturing local point cloud features, (ii) a novel feature aggregator that fuses features from multiple frames, (iii) an off-the-shelf backbone encoder for processing the aggregated features, and (iv) an efficient point decoder that generates per-point predictions directly from the encoder output. An overview of the \abbrev architecture is provided in \cref{fig:architecture}. In the following sections, we describe each component in detail.

\subsection{Local Feature Extraction}
\label{sec:lfe}

Directly processing dense LiDAR point clouds or aggregated features from multiple frames using a point-based backbone encoder can require substantial computational resources. As illustrated in \Cref{fig:latency}, the PTv3's runtime efficiency deteriorates with increasing input size, even when using voxel-grid downsampling.

To address this challenge and facilitate the use of additional frames, we employ a \emph{local feature extractor} (LFE) to capture point cloud features within a local neighbourhood. It reduces redundancies while preserving relevant local geometric information that will be used by the feature aggregator later.
To efficiently define the local neighborhood, we first voxelize the point cloud and assign each point to a voxel in the grid. An off-the-shelf voxel feature extractor (VFE) commonly used in 3D object detection~\cite{yan2018second} is employed to compute feature vectors for each voxel based on the points contained within it.

Although voxelization is applied, our method does not impose any structural assumptions on the resulting features like voxel grids. Instead, each voxel is treated as a new point, represented by a coordinate and an associated feature vector. Thus the resulting features form a new point cloud corresponding to the set of voxel features.


\subsection{Feature Aggregation}
\label{sec:feature-aggregation}

The \emph{feature aggregator} in \abbrev is designed to combine the LFE features from past frames with the current frame features. In order to retain the benefits of input concatenation as previously demonstrated in \cref{fig:motivation}, our feature aggregator aims to mimic the behavior of point cloud concatenation but in the feature space, allowing us to aggregate features in a way that maintains the spatial relationships between point clouds. To achieve this, our core idea is to formulate the feature space as a commutative monoid and treat the aggregator as a homomorphism mapping, thereby ensuring that the aggregation operation behaves similarly to point cloud concatenation.

Formally, let $P$ be the set of point clouds and $\oplus: P\times P\to P$ the concatenation operation. Since $\oplus$ is associative and commutative, and the empty point cloud $e\in P$ is an identity element, $(P, \oplus, e)$ thus forms a \emph{commutative monoid}. We further denote the LFE defined in the previous section as a function $f: P\to F$ that maps point clouds to a feature space $F$.

We first formulate $(F, \odot, \mathbf{0})$, the feature space $F$ equipped with the binary aggregator $\odot$ and the zero vector $\mathbf{0}$, to be a commutative monoid. Given features $x, y, z \in F$, the aggregator $\odot$ should satisfy the following properties:
\begin{align}
    &x\odot y = y\odot x, && \text{(Commutativity)} \label{eq:monoid-commutativity} \\
    &(x \odot y) \odot z = x \odot (y \odot z), && \text{(Associativity)} \label{eq:monoid-associativity} \\
    &x\odot \mathbf{0} = x. && \text{(Identity)} \label{eq:monoid-identity}
\end{align}
Since $(P, \oplus, e)$ and $(F, \odot, \mathrm{0})$ are now both commutative monoids, we can now treat the LFE $f$ as a \emph{homomorphism} between them. Specifically, for point clouds $u, v\in P$,
\begin{align}
    f(u\oplus v) &= f(u)\odot f(v), \label{eq:homomorphism-structure} \\
    f(e) &= \mathbf{0}. \label{eq:homomorphism-identity}
\end{align}

\begin{figure}
    \centering
    \includegraphics[width=0.9\linewidth]{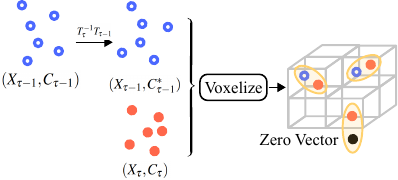}
    \caption{The feature association process (\cref{sec:feature-association}). Previous frame LFE features first undergo ego-motion correction, then are jointly voxelized with the current frame features to establish association. Features that cannot be associated are paired with the zero vector $\mathbf{0}$.}
    \label{fig:feature-association}
\end{figure}

In practical applications, the point cloud sequences are typically aggregated in the order that they are captured by the LiDAR sensor. While this suggests that $\odot$ may not need to be commutative or associative, we find these properties to be beneficial to the model's performance, as will be shown in the ablation study (\cref{sec:ablation}).

With the algebraic structure of the feature space defined, we next focus on how features from different frames are associated and fused during the aggregation process, and how the aforementioned properties can be enforced through architectural design and auxiliary losses.

\begin{figure*}
    \centering
    \includegraphics[width=0.9\linewidth]{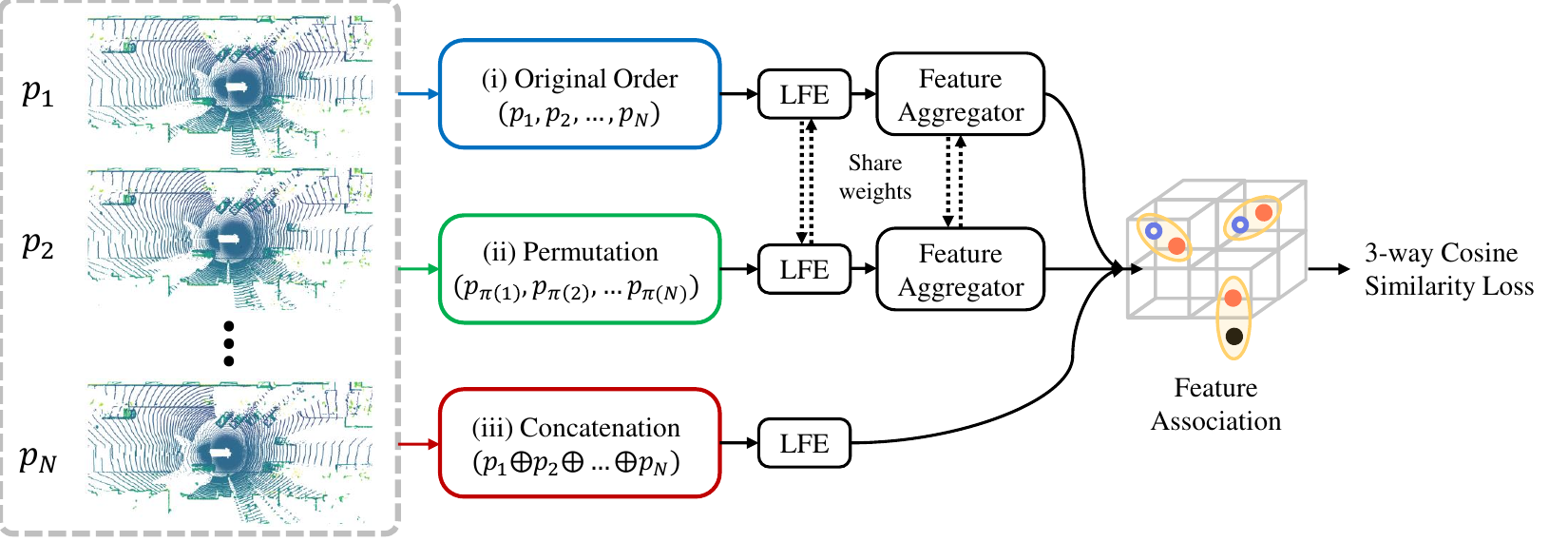}
    \caption{The proposed training scheme and auxiliary loss for the LFE and the feature aggregator.}
    \label{fig:auxiliary-loss}
\end{figure*}

\subsubsection{Feature Association} 
\label{sec:feature-association}
Recall from \cref{sec:lfe}, the LFE produces a set of feature vectors for each point cloud, where each feature vector is computed using the points in the corresponding voxel. To combine a set of features from previous frames $X_{\tau-1}\subseteq F$ with the current frame features $X_{\tau}\subseteq F$, we first need to associate the two sets of feature vectors before applying $\odot$ to each associated pair.

As depicted in \cref{fig:feature-association}, given features $X_{\tau-1}$ and  corresponding coordinates $C_{\tau-1} \subseteq \mathbb{R}^3$ from the previous frames, we first map the feature coordinates from the previous coordinate system $T_{\tau-1}\in \mathbb{SE}(3)$ into the current coordinate system $T_{\tau}$:
\begin{equation}
    C_{\tau-1}^* = \left\{ T_{\tau}^{-1}T_{\tau-1}\cdot c \ \vert\ c\in C_{\tau-1}\right\}.
\end{equation}

Afterward, feature vectors from the current frame $X_{\tau}$ and the past frames $X_{\tau-1}$ are assigned to voxels based on the current coordinates $C_{\tau}$ and the transformed coordinates $C_{\tau-1}^*$ respectively. We ensure that at most one feature vector from each set is assigned to the same voxel by using the same voxel size as VFE. The feature vectors belonging to the same voxel are associated and will be fused using $\odot$ in the following step.

Since the feature coordinates are generally not aligned with the voxel grid, we further append the offset from the voxel center to each feature vector to preserve fine-grained spatial information. For feature vectors that cannot find an association after voxelization, we set their associated vectors to be the zero vector $\mathbf{0}$, indicating that the point cloud in that region is empty $e$. Note that this trivially satisfies \cref{eq:homomorphism-identity}.

\subsubsection{Feature Aggregator}
\label{sec:feature-aggregator}
After the association step, the feature aggregator $\odot$ is applied to each pair of the associated feature vectors. We formulate the aggregator $\odot$ as follows:
\begin{equation}
    x\odot y = \begin{cases}
    x + y, & \text{if } x = \mathbf{0} \text{ or } y = \mathbf{0}, \\
    g\left( \frac{h(x,y)+h(y,x)}{2} \right), & \text{otherwise.}
  \end{cases}
  \label{eq:aggregator}
\end{equation}
$h(x, y): F\times F \to F$ is a binary operator that merges the associated feature vectors $x$ and $y$ via concatenation followed by an FCN; $g: F\to F$ is another FCN.

The first case in \cref{eq:aggregator} handles the situation where a feature vector cannot be associated, satisfying the identity property (\cref{eq:monoid-identity}). In the general case, we apply symmetrization to $h$ followed by another FCN $g$ to enforce commutativity (\cref{eq:monoid-commutativity}).

\subsubsection{Auxiliary Loss}
\label{sec:auxiliary-loss}
As the associativity (\cref{eq:monoid-associativity}) and the homomorphism (\cref{eq:homomorphism-structure}) properties cannot be strongly enforced via architectural design, we use an auxiliary loss to approximate them.

Given point clouds $p_1, p_2, p_3\in P$, we can rewrite \cref{eq:monoid-associativity,eq:homomorphism-structure} as the following:
\begin{equation}
\label{eq:homomorphism-associativity}
\begin{split}
    f(p_1\oplus p_2\oplus p_3) &=
    \big(f(p_1) \odot f(p_2)\big) \odot f(p_3) \\
    &= \left(f(p_{\pi(1)}) \odot f(p_{\pi(2)})\right) \odot f(p_{\pi(3)}),
\end{split}
\end{equation}
where $\pi$ is an arbitrary permutation. To represent the three terms in \cref{eq:homomorphism-associativity}, we process the point cloud sequences with three branches during training, as depicted in \cref{fig:auxiliary-loss}: (i) the sequence is aggregated in its original order, (ii) the sequence is aggregated under some random permutation, and (iii) the sequence is first concatenated at the input level, then processed by the LFE. The three sets of feature vectors are then associated with each other using the same technique as described in \cref{sec:feature-association}. For each associated triplet, we compute a three-way cosine similarity loss.

This auxiliary loss ensures that the LFE $f$ and feature aggregator $\odot$ approximate the desired algebraic properties, facilitating the network to learn a more robust and consistent feature aggregation process.

\subsection{Point Decoder}
\label{sec:point-decoder}


After applying an off-the-shelf backbone to the aggregated features, \abbrev employs an MLP-based \emph{point decoder} to generate per-point predictions. Unlike the U-Net-style decoders used in existing methods, our approach avoids upsampling the encoder features. This design serves two purposes: first, we aim to prevent unnecessary computation of features for points from previous frames. Second, since the LFE operates on a neighborhood of points and the aggregation occurs at the feature level, symmetrical upsampling back to the original input resolution is not feasible.

Instead, we propose a simple and efficient point decoder that avoids the upsampling process by directly using the encoder's output. As illustrated in \cref{fig:architecture}, for each point where a prediction is required, we query the encoder features at multiple scales using K-nearest neighbors (KNN) and append the relative offsets from the feature vector’s coordinates. The multi-scale features and offsets are then concatenated and processed by an FCN to generate the final segmentation prediction.

\begin{table*}[ht]
\centering
\caption{Results on the nuScenes validation split.}
\label{tab:results-nuscenes-val}
\resizebox{\textwidth}{!}{%
\begin{tabular}[p]{l|c|cccccccccccccccc}
\toprule
Method & mIoU & \rotatebox{90}{Barrier} & \rotatebox{90}{Bicycle} & \rotatebox{90}{Bus} & \rotatebox{90}{Car} & \rotatebox{90}{Const. veh.} & \rotatebox{90}{Motorcycle} & \rotatebox{90}{Pedestrian} & \rotatebox{90}{Traffic cone} & \rotatebox{90}{Trailer} & \rotatebox{90}{Truck} & \rotatebox{90}{Drivable} & \rotatebox{90}{Other flat} & \rotatebox{90}{Sidewalk} & \rotatebox{90}{Terrain} & \rotatebox{90}{Manmade} & \rotatebox{90}{Vegetation} \\ \midrule
Cylinder3D~\cite{zhu2021cylinder3d} & 76.1 & 76.4 & 40.3 & 91.2 & 93.8 & 51.3 & 78.0 & 78.9 & 64.9 & 62.1 & 84.4 & 96.8 & 71.6 & 76.4 & 75.4 & 90.5 & 87.4 \\
PVKD~\cite{hou2022pvkd} & 76.0 & 76.2 & 40.0 & 90.2 & 94.0 & 50.9 & 77.4 & 78.8 & 64.7 & 62.0 & 84.1 & 96.6 & 71.4 & 76.4 & 76.3 & 90.3 & 86.9 \\
RPVNet~\cite{xu2021rpvnet} & 77.6 & 78.2 & 43.4 & 92.7 & 93.2 & 49.0 & 85.7 & 80.5 & 66.0 & 66.9 & 84.0 & 96.9 & 73.5 & 75.9 & 76.0 & 90.6 & 88.9 \\
SphereFormer*~\cite{lai2023sphereformer} & 79.5 & 78.7 & 46.7 & 95.2 & 93.7 & 54.0 & 88.9 & 81.1 & 68.0 & 74.2 & 86.2 & \textbf{97.2} & 74.3 & 76.3 & 75.8 & 91.4 & 89.7 \\
PTv3*~\cite{wu2024ptv3} & 80.4 & \textbf{80.5} & 53.9 & \textbf{96.0} & 91.9 & 52.3 & 89.0 & 84.4 & 71.7 & 74.2 & 84.5 & \textbf{97.2} & \textbf{75.6} & 76.9 & 76.2 & 91.2 & 89.6 \\ 
PTv3 (10-frame)\textdagger & 81.3 & 79.3 & 61.0 & 95.9 & \textbf{95.0} & 50.7 & \textbf{93.0} & \textbf{87.3} & \textbf{74.5} & 68.8 & \textbf{89.2} & 97.0 & 74.1 & 76.5 & 76.3 & \textbf{91.6} & \textbf{90.4} \\ \midrule
MFSeg (ours) & 81.3 & 78.7 & 59.7 & 94.1 & 91.7 & 65.1 & 90.3 & 85.2 & 71.2 & 73.8 & 85.6 & 96.9 & 74.4 & 76.5 & 76.5 & 91.2 & 89.8 \\
w/ TTA* & \textbf{82.5} & 79.4 & \textbf{62.1} & 94.7 & 92.5 & \textbf{67.1} & 92.2 & 86.7 & 73.4 & \textbf{75.1} & 86.5 & 97.1 & \textbf{75.6} & \textbf{77.8} & \textbf{77.5} & \textbf{91.6} & \textbf{90.4} \\ \midrule
\multicolumn{18}{c}{* indicates the method uses test-time augmentation (TTA). \textdagger\ results are reproduced by us using the official codebase.}
\end{tabular}%
}
\end{table*}

\section{Experimental Setup}
\subsection{Datasets}
To evaluate the effectiveness and efficiency of \abbrev, we perform experiments on two large scale semantic segmentation benchmarks: nuScenes~\cite{caesar2020nuscenes} and Waymo~\cite{sun2020waymo}.

The nuScenes dataset is collected in the Boston and Singapore areas using a 32-beam LiDAR. It consists of 1000 sequences with 850 in the training and validation splits, and 150 reserved for testing. Due to the relatively sparse nature of the point clouds in this dataset, with an average of 37k points per frame, 3D perception tasks can often benefit from input concatenation across longer sequences.

The Waymo dataset is collected in the San Francisco, Mountain View, and Phoenix regions using a more complex 5-sensor setup, with a primary 64-beam LiDAR. Consequently, the point clouds in Waymo are much denser, averaging 177k points per frame. This density poses a significantly greater computational challenge compared to nuScenes, particularly with multi-frame inputs. 


\subsection{Implementation Details}
Our implementation is built on the official codebase of PTv3~\footnote{https://github.com/Pointcept/Pointcept}, utilizing the PTv3's encoder as our backbone. The LFE module consists of four linear layers with channel dimensions of $[128, 128, 128, 128]$. The feature aggregator is composed of two FCNs, $h$ and $g$, as described in \cref{sec:feature-aggregator}. Both $h$ and $g$ are three-layer FCNs with a hidden channel dimension of 128. Following a common practice in input concatenation, during feature aggregation, we append timestamps to the feature vectors to help the model distinguish current and past frame features. Finally, the point decoder is an FCN with hidden channel sizes $[1024, 512, 256, 128, 64]$. For nuScenes models, we aggregate a maximum of 10 frames, and for Waymo models, we aggregate a maximum of 5 frames. 

We employ a two-stage training strategy. In the first stage, we train the model with input concatenation, bypassing the feature aggregator (branch (iii) in \cref{fig:auxiliary-loss}). This helps initialize the LFE and the other network components. For this stage, we adopt the same training parameters and data augmentations used by PTv3. For both the nuScenes and Waymo datasets, the model is trained for 50 epochs following a one-cycle learning rate schedule with a maximum learning rate of 0.002. In the second stage, we fine-tune the model using the auxiliary loss scheme detailed in \cref{sec:auxiliary-loss}. The model is trained for an additional 20 epochs with a maximum learning rate of 0.001.

\begin{table*}[ht]
\centering
\caption{Results on the Waymo validation split.}
\label{tab:results-waymo-val}
\setlength\tabcolsep{0.4em}
\resizebox{\textwidth}{!}{%
\begin{tabular}{l|c|cccccccccccccccccccccc}
\toprule
Method & mIoU & \rotatebox{90}{Car} & \rotatebox{90}{Truck} & \rotatebox{90}{Bus} & \rotatebox{90}{Other veh.} & \rotatebox{90}{Motorcyclist} & \rotatebox{90}{Bicyclist} & \rotatebox{90}{Pedestrian} & \rotatebox{90}{Sign} & \rotatebox{90}{Traffic light} & \rotatebox{90}{Pole} & \rotatebox{90}{Con. cone} & \rotatebox{90}{Bicycle} & \rotatebox{90}{Motorcycle} & \rotatebox{90}{Building} & \rotatebox{90}{Vegetation} & \rotatebox{90}{Tree trunk} & \rotatebox{90}{Curb} & \rotatebox{90}{Road} & \rotatebox{90}{Lane marker} & \rotatebox{90}{Other gro.} & \rotatebox{90}{Walkable} & \rotatebox{90}{Sidewalk} \\ \midrule
SparseUNet~\cite{graham2018spunet} & 66.6 & 94.4 & 59.8 & 85.1 & 37.8 & 2.2 & 69.1 & 89.3 & 73.4 & 40.4 & 74.8 & 57.3 & 66.6 & 75.2 & 95.5 & 91.3 & 67.0 & 68.1 & 92.3 & 41.7 & 30.1 & 79.0 & 75.6 \\
SphereFormer~\cite{lai2023sphereformer} & 69.9 & 94.5 & 61.6 & 87.7 & 40.2 & 0.9 & 69.7 & 90.2 & 73.9 & \textbf{41.8} & 77.2 & 65.4 & 71.9 & 83.7 & 95.9 & 91.7 & 68.4 & 69.8 & 93.3 & 53.9 & 47.9 & 80.8 & 77.2 \\
PTv3*~\cite{wu2024ptv3} & 71.2 & \textbf{94.8} & \textbf{63.6} & \textbf{89.0} & 34.7 & \textbf{12.2} & 73.2 & 90.6 & 71.6 & 31.3 & 77.8 & \textbf{75.9} & 75.6 & 87.3 & 96.0 & 91.8 & 68.1 & 71.2 & 93.5 & 56.6 & 51.3 & 81.6 & 78.4 \\ \midrule
MFSeg (ours) & 72.1 & 94.5 & 62.2 & 86.1 & 43.7 & 8.6 & 79.0 & 91.1 & 75.1 & 33.5 & 80.4 & 72.2 & 78.3 & 90.0 & 96.1 & 92.0 & 70.0 & 72.2 & 93.4 & 56.9 & 51.2 & 81.2 & 78.0 \\
w/ TTA* & \textbf{72.7} & 94.6 & 63.0 & 86.2 & \textbf{44.1} & 4.5 & \textbf{81.4} & \textbf{91.6} & \textbf{76.0} & 34.9 & \textbf{81.4} & 73.9 & \textbf{79.4} & \textbf{92.3} & \textbf{96.3} & \textbf{92.2} & \textbf{70.7} & \textbf{73.1} & \textbf{93.6} & \textbf{57.6} & \textbf{52.0} & \textbf{81.7} & \textbf{78.8} \\ \midrule
\multicolumn{24}{c}{* indicates the method uses test-time augmentation (TTA).}
\end{tabular}%
}
\end{table*}

\section{Results}
\subsection{Overall Performance}
The results on the nuScenes dataset and Waymo dataset are presented in \cref{tab:results-nuscenes-val} and \cref{tab:results-waymo-val} respectively. To provide a fair comparison with methods that utilize test-time augmentation (TTA), we report results both with and without TTA, although TTA is not practical for real-time applications.

\begin{figure}
    \centering
    \includegraphics[width=0.40\linewidth]{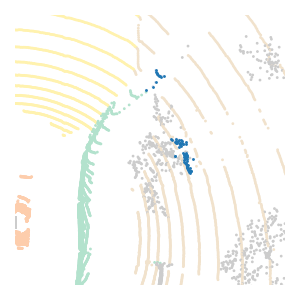}
    \includegraphics[width=0.40\linewidth]{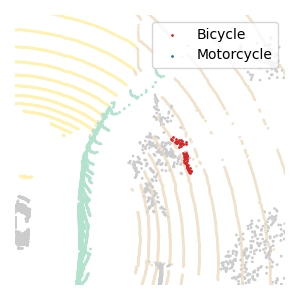}
    \caption{Segmentation results from PTv3 (left) and MFSeg (right). PTv3 mis-classifies a partially occluded bicycle along with nearby background points as motorcycle (blue), while MFSeg accurately identifies the points belonging to the bicycle (red).}
    \vspace{-1em}
    \label{fig:qualitative-bicycle}
\end{figure}

MFSeg outperforms competing single-frame methods across both datasets, even without the use of TTA. On the nuScenes dataset, MFSeg achieves a mIoU of 81.3\% without TTA and 82.5\% with TTA, surpassing the performance of the SOTA method PTv3 with single-frame input, while matching its performance with 10-frame concatenated point clouds. Notably, compared to single-frame methods, MFSeg improves the performance of vulnerable road users, including pedestrian, bicycle and motorcycle, which are crucial in practical applications. A qualitative example is shown in \cref{fig:qualitative-bicycle}.

On the Waymo dataset, MFSeg also outperforms single-frame SOTA PTv3 in terms of mIoU, achieving 72.1\% without TTA and 72.7\% with TTA. Similarly, we observe substantial improvements in the performance of bicycle, bicyclist, and motorcycle, with a moderate improvement in pedestrian. 

Our results demonstrate that MFSeg can effectively utilize sequential point cloud data, providing benefits similar to point cloud concatenation (\cref{fig:motivation}) without the associated computational cost. Moreover, MFSeg's ability to perform well without relying on TTA highlights its potential for real-time deployment.

\subsection{Inference Latency}
To demonstrate that MFSeg can process multi-frame point clouds more efficiently than existing point-based methods with concatenated point clouds, we compare its inference time with SOTA method PTv3 across a range of input sequence lengths.

Although PTv3 is a point-based method, in practice, its input point cloud is downsampled using a voxel-grid. To highlight the effect of input complexity on the latency of point-based methods without downsampling, we evaluate PTv3 using a small voxel size of 0.01 $\mathrm{m}^3$ to minimize the downsampling effects. This setup demonstrates the computational requirements of processing dense point clouds without simplification. For comparison, we refer to the version of PTv3 that employs voxel-grid downsampling as PTv3-ds.

\begin{table*}[ht]
\centering
\caption{Ablation study on the NuScenes validation split. All models are evaluated without TTA.}
\label{tab:ablation-nuscenes}
\resizebox{0.65\textwidth}{!}{%
\begin{tabular}{c|cccccc|c}
\toprule
 & LFE & \begin{tabular}[c]{@{}c@{}}Point\\ Decoder\end{tabular} & \begin{tabular}[c]{@{}c@{}}Input\\ Concatenation\end{tabular} & \begin{tabular}[c]{@{}c@{}}Feature\\ Aggregation\end{tabular} & \begin{tabular}[c]{@{}c@{}}Associativity\\ Loss\end{tabular} & \begin{tabular}[c]{@{}c@{}}Homomorphism\\ Loss\end{tabular} & mIoU \\ \midrule
A & - & - & - & - & - & - & 78.6 \\
B & \checkmark & \checkmark & - & - & - & - & 79.2 \\
C & \checkmark & \checkmark & \checkmark & - & - & - & 81.7 \\
D & \checkmark & \checkmark & - & \checkmark & - & - & 79.8 \\
E & \checkmark & \checkmark & - & \checkmark & \checkmark & - & 80.6 \\
F & \checkmark & \checkmark & - & \checkmark & \checkmark & \checkmark & 81.3 \\ \bottomrule
\end{tabular}%
}
\vspace{-1em}
\end{table*}

We measure the latency on the validation split using a single NVIDIA A6000 GPU with minimum additional system load. Data loading time is excluded from the measurement. The results are shown in \cref{fig:latency}. In both datasets, we observe that PTv3, when processing concatenated point clouds with minimal downsampling, scales poorly as the number of frames increases, making it unsuitable for online settings. While PTv3-ds has a much lower latency thanks to input downsampling, it is unable to produce predictions for every point in the point cloud without label interpolation or multiple forward passes. In contrast, \abbrev achieves lower latency and scales more efficiently as the number of frames increases, indicated by the gentler slope, while delivering better performance compared to input concatenation.

\begin{figure}
    \centering
    \includegraphics[width=0.9\linewidth]{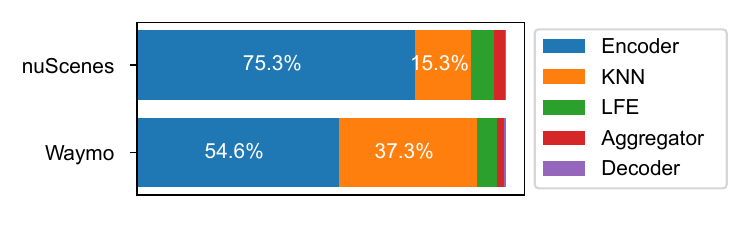}
    \caption{A breakdown of the latency of each component in MFSeg.}
    \vspace{-1em}
    \label{fig:latency-breakdown}
\end{figure}

\cref{fig:latency-breakdown} shows a breakdown of the latency of each component of MFSeg. In both datasets, we observe that the proposed LFE layer and point decoder introduce minimum computational overhead. However, as the input density increases, particularly in the Waymo dataset, the cost of the KNN operation rises significantly. Efficiently extracting multi-scale features from the unstructured features of a point-based backbone remains a computational challenge, and we believe future work should focus on addressing this issue.

\subsection{Ablation Study}
\label{sec:ablation}
We conduct an ablation study on the nuScenes dataset to illustrate the impact of each proposed component. The results are presented in \cref{tab:ablation-nuscenes}.

First, in row A and B, we investigate the effects of LFE and the point decoder in the single-frame setting. Note that row A represents a baseline PTv3 without our proposed modifications, and the feature aggregator is bypassed in this setup. The results show that both models achieve comparable performance, indicating that the LFE and point decoder maintain the SOTA performance of PTv3, despite being simpler in design.

Next, we explore the impact of input concatenation and feature-level aggregation in multi-frame settings. Comparing row C with rows A and B, we observe that multi-frame point clouds can enhance segmentation performance when using a point-based backbone. However, in row D, where feature aggregation is applied without regularization, the model can no longer achieve the full benefits of input concatenation.

By introducing auxiliary losses to enforce the desired algebraic properties in the feature space—-specifically associativity (row E) and homomorphism (row F)—-we observe that our method matches the performance of input concatenation (row C). In row E, we disable the third branch depicted in \cref{fig:auxiliary-loss}, and the feature association and cosine similarity is only computed between the original sequence and permuted sequence, thus only enforcing \cref{eq:monoid-associativity}. Row F represents our full approach with all the proposed components. These results validate the effectiveness of our proposed method.

\section{Limitations And Future Work}
\abbrev has two primary limitations. First, although our feature aggregator effectively mimics the behavior of point cloud concatenation, it operates at the local neighborhood level. As a result, the aggregated features may lack the contextual information available in the deeper layers of the network. To fully take advantage of the global context from later stages of the network, modifications to the current architecture might be necessary to integrate both local and global features.

Second, although the design of the point decoder is simple and imposes minimal computational cost (\cref{fig:latency-breakdown}), the KNN operation that is required to gather multi-scale features from the backbone encoder is computationally expensive. This operation can become a bottleneck in real-time applications. To further improve the efficiency of our approach, replacing the KNN with a voxel-based association technique, such as the one used in \cref{sec:feature-association}, could improve the scalability of MFSeg.

\section{Conclusion}
In this paper, we presented MFSeg, an efficient framework for multi-frame 3D semantic segmentation. By aggregating point cloud sequences at the feature level and regularizing the feature extraction and aggregation process using a novel formulation, MFSeg is able to reduce computational overhead while maintaining high accuracy. Experiments on the nuScenes and Waymo datasets demonstrate that MFSeg outperforms existing methods while achieving lower inference latency than the comparable point-based method in the multi-frame setting.

\newpage
\bibliographystyle{ieee_fullname}
\bibliography{main}

\begin{thebibliography}{10}\itemsep=-1pt

\bibitem{caesar2020nuscenes}
Holger Caesar, Varun Bankiti, Alex~H. Lang, Sourabh Vora, Venice~Erin Liong, Qiang Xu, Anush Krishnan, Yu Pan, Giancarlo Baldan, and Oscar Beijbom.
\newblock {nuScenes}: A multimodal dataset for autonomous driving.
\newblock In {\em IEEE/CVF Conf. on Computer Vision and Pattern Recognition (CVPR)}, June 2020.

\bibitem{chen2022mppnet}
Xuesong Chen, Shaoshuai Shi, Benjin Zhu, Ka~Chun Cheung, Hang Xu, and Hongsheng Li.
\newblock {MPPNet}: Multi-frame feature intertwining with proxy points for {3D} temporal object detection.
\newblock In {\em European Conf. on Computer Vision (ECCV)}, pages 680--697, Cham, 2022. Springer Nature Switzerland.

\bibitem{choy2019minkunet}
Christopher Choy, JunYoung Gwak, and Silvio Savarese.
\newblock 4d spatio-temporal convnets: Minkowski convolutional neural networks.
\newblock In {\em IEEE/CVF Conf. on Computer Vision and Pattern Recognition (CVPR)}, pages 3075--3084, 2019.

\bibitem{dao2022flashattention}
Tri Dao, Dan Fu, Stefano Ermon, Atri Rudra, and Christopher R{\'e}.
\newblock Flashattention: Fast and memory-efficient exact attention with io-awareness.
\newblock {\em NeurIPS}, 35:16344--16359, 2022.

\bibitem{graham2018spunet}
Benjamin Graham, Martin Engelcke, and Laurens Van Der~Maaten.
\newblock 3d semantic segmentation with submanifold sparse convolutional networks.
\newblock In {\em IEEE/CVF Conf. on Computer Vision and Pattern Recognition (CVPR)}, pages 9224--9232, 2018.

\bibitem{graham2017spconv}
Benjamin Graham and Laurens Van~der Maaten.
\newblock Submanifold sparse convolutional networks.
\newblock {\em arXiv preprint arXiv:1706.01307}, 2017.

\bibitem{hou2022pvkd}
Yuenan Hou, Xinge Zhu, Yuexin Ma, Chen~Change Loy, and Yikang Li.
\newblock Point-to-voxel knowledge distillation for lidar semantic segmentation.
\newblock In {\em Proceedings of the IEEE/CVF conference on computer vision and pattern recognition}, pages 8479--8488, 2022.

\bibitem{huang2024soap}
Chengjie Huang, Vahdat Abdelzad, Sean Sedwards, and Krzysztof Czarnecki.
\newblock {SOAP}: Cross-sensor domain adaptation for {3D} object detection using {S}tationary {O}bject {A}ggregation {P}seudo-labelling.
\newblock In {\em IEEE/CVF Winter Conf. on Applications of Computer Vision (WACV)}, 2024.

\bibitem{lai2023sphereformer}
Xin Lai, Yukang Chen, Fanbin Lu, Jianhui Liu, and Jiaya Jia.
\newblock Spherical transformer for lidar-based 3d recognition.
\newblock In {\em IEEE/CVF Conf. on Computer Vision and Pattern Recognition (CVPR)}, pages 17545--17555, 2023.

\bibitem{li2023mseg3d}
Jiale Li, Hang Dai, Hao Han, and Yong Ding.
\newblock Mseg3d: Multi-modal 3d semantic segmentation for autonomous driving.
\newblock In {\em IEEE/CVF Conf. on Computer Vision and Pattern Recognition (CVPR)}, pages 21694--21704, 2023.

\bibitem{liu2023uniseg}
Youquan Liu, Runnan Chen, Xin Li, Lingdong Kong, Yuchen Yang, Zhaoyang Xia, Yeqi Bai, Xinge Zhu, Yuexin Ma, Yikang Li, et~al.
\newblock Uniseg: A unified multi-modal lidar segmentation network and the openpcseg codebase.
\newblock In {\em IEEE Int. Conf. on Computer Vision (ICCV)}, pages 21662--21673, 2023.

\bibitem{liu2019spvnas}
Zhijian Liu, Haotian Tang, Yujun Lin, and Song Han.
\newblock Point-voxel cnn for efficient 3d deep learning.
\newblock {\em NeurIPS}, 32, 2019.

\bibitem{luo2020latticenet}
Xiaotong Luo, Yuan Xie, Yulun Zhang, Yanyun Qu, Cuihua Li, and Yun Fu.
\newblock Latticenet: Towards lightweight image super-resolution with lattice block.
\newblock In {\em European Conf. on Computer Vision (ECCV)}, pages 272--289. Springer, 2020.

\bibitem{milioto2019rangenet++}
Andres Milioto, Ignacio Vizzo, Jens Behley, and Cyrill Stachniss.
\newblock Rangenet++: Fast and accurate lidar semantic segmentation.
\newblock In {\em IEEE/RSJ international conference on intelligent robots and systems (IROS)}, pages 4213--4220. IEEE, 2019.

\bibitem{qi2017pointnet}
Charles~R Qi, Hao Su, Kaichun Mo, and Leonidas~J Guibas.
\newblock Pointnet: Deep learning on point sets for 3d classification and segmentation.
\newblock In {\em IEEE/CVF Conf. on Computer Vision and Pattern Recognition (CVPR)}, pages 652--660, 2017.

\bibitem{qi2017pointnet++}
Charles~Ruizhongtai Qi, Li Yi, Hao Su, and Leonidas~J Guibas.
\newblock Pointnet++: Deep hierarchical feature learning on point sets in a metric space.
\newblock {\em NeurIPS}, 30, 2017.

\bibitem{schutt2022temporallatticenet}
Peer Schutt, Radu~Alexandru Rosu, and Sven Behnke.
\newblock Abstract flow for temporal semantic segmentation on the permutohedral lattice.
\newblock In {\em Int. Conf. on Robotics and Automation (ICRA)}, pages 5139--5145. IEEE, 2022.

\bibitem{shi2020spsequencenet}
Hanyu Shi, Guosheng Lin, Hao Wang, Tzu-Yi Hung, and Zhenhua Wang.
\newblock Spsequencenet: Semantic segmentation network on 4d point clouds.
\newblock In {\em Proceedings of the IEEE/CVF conference on computer vision and pattern recognition}, pages 4574--4583, 2020.

\bibitem{sun2020waymo}
Pei Sun, Henrik Kretzschmar, Xerxes Dotiwalla, Aurelien Chouard, Vijaysai Patnaik, Paul Tsui, James Guo, Yin Zhou, Yuning Chai, Benjamin Caine, Vijay Vasudevan, Wei Han, Jiquan Ngiam, Hang Zhao, Aleksei Timofeev, Scott Ettinger, Maxim Krivokon, Amy Gao, Aditya Joshi, Yu Zhang, Jonathon Shlens, Zhifeng Chen, and Dragomir Anguelov.
\newblock Scalability in perception for autonomous driving: {W}aymo {O}pen {D}ataset.
\newblock In {\em IEEE/CVF Conf. on Computer Vision and Pattern Recognition (CVPR)}, June 2020.

\bibitem{wang2022meta}
Song Wang, Jianke Zhu, and Ruixiang Zhang.
\newblock Meta-rangeseg: Lidar sequence semantic segmentation using multiple feature aggregation.
\newblock {\em IEEE Robotics and Automation Letters}, 7(4):9739--9746, 2022.

\bibitem{wu2018squeezeseg}
Bichen Wu, Alvin Wan, Xiangyu Yue, and Kurt Keutzer.
\newblock Squeezeseg: Convolutional neural nets with recurrent crf for real-time road-object segmentation from 3d lidar point cloud.
\newblock In {\em Int. Conf. on Robotics and Automation (ICRA)}, pages 1887--1893. IEEE, 2018.

\bibitem{wu2019squeezesegv2}
Bichen Wu, Xuanyu Zhou, Sicheng Zhao, Xiangyu Yue, and Kurt Keutzer.
\newblock Squeezesegv2: Improved model structure and unsupervised domain adaptation for road-object segmentation from a lidar point cloud.
\newblock In {\em Int. Conf. on Robotics and Automation (ICRA)}, pages 4376--4382. IEEE, 2019.

\bibitem{wu2024ptv3}
Xiaoyang Wu, Li Jiang, Peng-Shuai Wang, Zhijian Liu, Xihui Liu, Yu Qiao, Wanli Ouyang, Tong He, and Hengshuang Zhao.
\newblock Point transformer v3: Simpler faster stronger.
\newblock In {\em IEEE/CVF Conf. on Computer Vision and Pattern Recognition (CVPR)}, pages 4840--4851, 2024.

\bibitem{wu2022ptv2}
Xiaoyang Wu, Yixing Lao, Li Jiang, Xihui Liu, and Hengshuang Zhao.
\newblock Point transformer v2: Grouped vector attention and partition-based pooling.
\newblock {\em NeurIPS}, 35:33330--33342, 2022.

\bibitem{xu2020squeezesegv3}
Chenfeng Xu, Bichen Wu, Zining Wang, Wei Zhan, Peter Vajda, Kurt Keutzer, and Masayoshi Tomizuka.
\newblock Squeezesegv3: Spatially-adaptive convolution for efficient point-cloud segmentation.
\newblock In {\em European Conf. on Computer Vision (ECCV)}, pages 1--19. Springer, 2020.

\bibitem{xu2021rpvnet}
Jianyun Xu, Ruixiang Zhang, Jian Dou, Yushi Zhu, Jie Sun, and Shiliang Pu.
\newblock Rpvnet: A deep and efficient range-point-voxel fusion network for lidar point cloud segmentation.
\newblock In {\em IEEE Int. Conf. on Computer Vision (ICCV)}, pages 16024--16033, 2021.

\bibitem{yan20222dpass}
Xu Yan, Jiantao Gao, Chaoda Zheng, Chao Zheng, Ruimao Zhang, Shuguang Cui, and Zhen Li.
\newblock 2dpass: 2d priors assisted semantic segmentation on lidar point clouds.
\newblock In {\em European Conf. on Computer Vision (ECCV)}, pages 677--695. Springer, 2022.

\bibitem{yan2018second}
Yan Yan, Yuxing Mao, and Bo Li.
\newblock {SECOND}: Sparsely embedded convolutional detection.
\newblock {\em Sensors}, 18(10):3337, 2018.

\bibitem{yang20213dman}
Zetong Yang, Yin Zhou, Zhifeng Chen, and Jiquan Ngiam.
\newblock {3D-MAN}: {3D} multi-frame attention network for object detection.
\newblock In {\em IEEE/CVF Conf. on Computer Vision and Pattern Recognition (CVPR)}, pages 1863--1872, June 2021.

\bibitem{zhang2020polarnet}
Yang Zhang, Zixiang Zhou, Philip David, Xiangyu Yue, Zerong Xi, Boqing Gong, and Hassan Foroosh.
\newblock Polarnet: An improved grid representation for online lidar point clouds semantic segmentation.
\newblock In {\em IEEE/CVF Conf. on Computer Vision and Pattern Recognition (CVPR)}, pages 9601--9610, 2020.

\bibitem{zhao2021pt}
Hengshuang Zhao, Li Jiang, Jiaya Jia, Philip~HS Torr, and Vladlen Koltun.
\newblock Point transformer.
\newblock In {\em IEEE Int. Conf. on Computer Vision (ICCV)}, pages 16259--16268, 2021.

\bibitem{zhu2021cylinder3d}
Xinge Zhu, Hui Zhou, Tai Wang, Fangzhou Hong, Yuexin Ma, Wei Li, Hongsheng Li, and Dahua Lin.
\newblock Cylindrical and asymmetrical 3d convolution networks for lidar segmentation.
\newblock In {\em IEEE/CVF Conf. on Computer Vision and Pattern Recognition (CVPR)}, pages 9939--9948, 2021.

\end{thebibliography}

\end{document}